\documentclass[11pt]{article}

\usepackage[preprint]{acl}

\usepackage{times}
\usepackage{latexsym}

\usepackage[T1]{fontenc}
\usepackage[utf8]{inputenc}

\usepackage{microtype}

\usepackage{inconsolata}

\usepackage{graphicx}

\usepackage{amsfonts,amsmath,amssymb}
\usepackage{booktabs}
\usepackage{enumitem}

\title{Evaluating Geographic Lexical Knowledge in Large Language Models Across 21 Spanish-Speaking Countries}

\author{%
  Yoshifumi Kawasaki\\
  The University of Tokyo\\
  \texttt{ykawasaki@g.ecc.u-tokyo.ac.jp}
  }

\begin{document}
\maketitle

\begin{abstract}
Large language models (LLMs) increasingly serve users across regional varieties, yet their knowledge of geographic lexical variation remains insufficiently evaluated.
We investigate country-conditioned lexical usage judgments in Spanish using an expert-curated database covering 934 lexical items across 21 Spanish-speaking countries.
Treating ten proprietary and open-weight LLMs as virtual informants, we evaluate two complementary question formats: Yes–No question format (YNQF) and multiple-choice question format (MCQF).
The evaluation comprises 10,164 YNQF questions and 16,637 MCQF questions, assessed using F1 and an adjusted Jaccard coefficient, respectively.
Results reveal substantial differences across models, countries, and formats.
Spain receives the highest country-level score for eight models in YNQF and nine in MCQF; Chile receives the lowest MCQF score for every model.
Regression analyses suggest that digital resource volume is generally more strongly associated with YNQF performance, whereas lexical distance from Spain is more strongly associated with MCQF performance.
These findings provide a fine-grained diagnostic of uneven regional lexical knowledge and highlight the importance of complementary elicitation formats and evaluation metrics.
The benchmark measures explicit usage judgments, motivating separate assessment of region-appropriate generation.
\end{abstract}

\section{Introduction}
\label{sec:introduction}
Spanish is spoken as a first language by more than 520 million people~\cite{institutocervantes-2025-anuario} and exhibits substantial geographic lexical variation across the Spanish-speaking world.
For many everyday concepts, multiple lexical variants coexist, with their distribution shaped by historical, regional, and sociolinguistic factors.
A well-known example is a variation in referring to a vehicle equivalent to English \textit{car}; they mainly use \textit{coche} in Spain, whereas \textit{carro} is preferred in many Latin American countries~\cite{ueda-morenofernandez-2016-varilex}.
Documenting and modeling this variation has long been a central concern of Spanish dialectology~\cite{zamoravicente-1968-dialectologia, lipski-1994-espanol, alvar-1996-manual-america, alvar-1996-manual-espana, fernandez-ordonez-2011-lengua, gonçalves-sanchez-2014-crowdsourcing, ueda-morenofernandez-2016-varilex, morenofernandez-2020-lengua, morenofernandez-2020-variedades, morenofernandez-rocio-2023-dialectologia, ueda-2023-dialectologia}.

Large Language Models (LLMs) are now widely used in downstream applications that interact directly with speakers of different regional varieties.
For dialect-sensitive applications, LLMs should understand diverse variants while selecting expressions appropriate to the intended users~\cite{lakew-etal-2018-neural, sun-etal-2023-dialect}.
However, these models are trained primarily on large-scale web data, whose composition is estimated to be uneven across regions and to privilege standardized or hegemonic varieties in detriment of dialect fairness~\cite{kantharuban-etal-2023-quantifying, munozbasols-et-al-2024-sesgo, lin-etal-2025-assessing}.
As a result, concerns have been raised that LLMs may inadequately represent dialectal diversity, thereby producing \emph{Digital Linguistic Bias} (DLB)~\cite{munozbasols-et-al-2024-sesgo}.
This may undermine real-world linguistic diversity in the digital sphere, causing linguistic homogenization through LLMs.

LLMs presumably encode dialectal information implicitly and can be prompted to exhibit region-specific lexical or stylistic features.
However, systematic evaluation of this knowledge remains limited, particularly for languages such as Spanish~\cite{mayorrocher-et-al-2025-igual}, where geographic variation is rich and well documented~\cite{morenofernandez-2020-lengua, morenofernandez-2020-variedades, morenofernandez-rocio-2023-dialectologia}.
Therefore, clarifying the extent to which LLMs can handle Spanish varieties is essential for developing LLMs that represent its linguistic and cultural diversity~\cite{grandury-et-al-2025-leaderboard}.

\begin{figure}[tb]
  \includegraphics[width=\columnwidth]{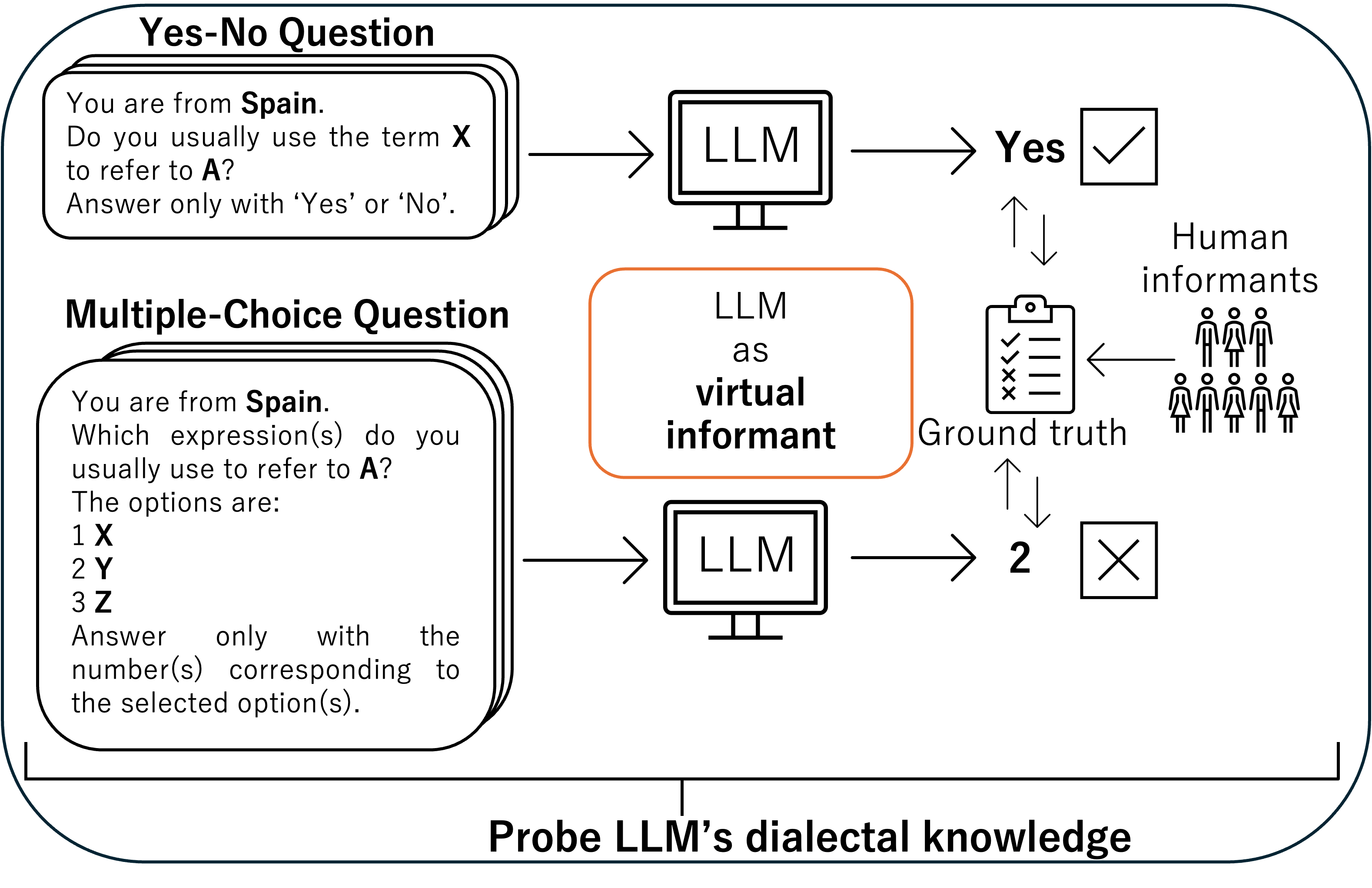}
  \caption{Schematic overview of the study. The questions are translated into English for clarity.}
  \label{fig:overview}
\end{figure}

In this paper, we provide a large-scale evaluation of geographic lexical variation in Spanish as represented in LLMs (Figure~\ref{fig:overview}).
We treat LLMs as \emph{virtual informants} and probe their knowledge of regional lexical usage using two survey-style question formats.
Thus, we provide a controlled diagnostic that complements evaluation of spontaneous text generation.

This work makes three main contributions:
\begin{itemize}
    \item \textbf{Large-scale evaluation of Spanish lexical variation:}
    We evaluate ten LLMs using an expert-curated resource covering 934 lexical items across 21 countries, providing a detailed assessment of country-conditioned lexical usage judgments.
    \item \textbf{Analysis of evaluation design:}
    We compare Yes--No and multiple-choice question formats, showing that elicitation choices affect how geographic performance disparities are interpreted.
    \item \textbf{Characterization of geographic disparities:}
    We identify shared and model-specific country-level patterns and examine their associations with digital resource availability and lexical distance, revealing that these relationships differ across evaluation formats.
\end{itemize}

The remainder of the paper is organized as follows: Section~\ref{sec:related_work} reviews related work, Section~\ref{sec:methods} describes the methodology, Section~\ref{sec:experiment} presents the experiments and results,  Section~\ref{sec:discussion} discusses the findings, and Section~\ref{sec:conclusions} concludes the paper.

\section{Related Work}
\label{sec:related_work}
Research on dialectal variation in language models has documented uneven performance across varieties in both targeted evaluations and downstream tasks.
\citet{kantharuban-etal-2023-quantifying} identify dialect gaps in machine translation and speech recognition across languages, while \citet{lin-etal-2025-assessing} report disparities between standardized English and African American Vernacular English in reasoning tasks.
These findings raise concerns about dialect fairness~\cite{davidson-etal-2019-racial, gallegos-etal-2024-bias} and \textit{Digital Linguistic Bias}, including unequal representation of varieties within Spanish~\cite{munozbasols-et-al-2024-sesgo}.

For Spanish, recent work has developed resources and evaluation frameworks that explicitly address regional diversity.
\citet{espanabonet-barroncedeno-2024-elote} introduce \textit{CEREAL}, a corpus of Spanish varieties, and investigate regional differences through variety classification and word embeddings.
\citet{grandury-et-al-2025-leaderboard} introduce \textit{La Leaderboard}, which combines evaluation datasets spanning Spanish varieties and other languages of Spain.
At the downstream level, \citet{bengoetxea-etal-2025-lost} examine natural-language inference in Spanish geographical varieties, observing performance degradation under linguistic variation.
These studies motivate geographically sensitive evaluation, while leaving room for more systematic analysis of specific linguistic dimensions.

The closest study to ours is \citet{mayorrocher-et-al-2025-igual}, which evaluates nine LLMs using 30 multiple-choice questions: 20 on morphosyntactic variation and 10 on lexical variation. 
Their questions draw on dictionaries, corpora, and \textit{VARILEX}~\cite{ueda-morenofernandez-2016-varilex}, and performance is assessed across seven conventional dialectal areas.
They find that Peninsular Spanish and La Plata River varieties are better distinguished, whereas Andean and Mexican \& Central American varieties are more difficult for the models to recognize.
They associate these differences with the amount of digital resources available across areas, reporting a strong positive correlation.
However, the relationship between resource availability and dialect performance warrants further examination: \citet{kantharuban-etal-2023-quantifying} find that data-related effects vary across models and languages.

Building on \citet{mayorrocher-et-al-2025-igual}, our study expands the lexical scope and geographic resolution of dialectal evaluation through an expert-curated resource covering 934 lexical items across 21 countries.
By combining Yes--No and multiple-choice questions, we provide a fine-grained assessment of country-conditioned lexical usage judgments in LLMs.

\section{Methods}
\label{sec:methods}
We evaluate the extent to which LLMs capture geographic lexical variation in Spanish by treating them as \emph{virtual informants} and eliciting dialectal judgments through structured question–answering tasks.
This section describes the reference dataset, the question formats used to probe the models, and the evaluation metrics.

\subsection{Reference Dataset: VARILEX}
\label{subsec:varilex}
As a gold-standard reference, we work with \textit{VARILEX (Variación léxica del español en el mundo)}, a large-scale, expert-curated database documenting geographic lexical variation across the Spanish-speaking world~\cite{ueda-morenofernandez-2016-varilex}.
We rely on its revised version\footnote{\url{https://h-ueda.sakura.ne.jp/varilex-r/}}, in which local linguists annotated the usage of lexical variants at the country level using a binary scheme: ``+'' denotes predominant usage, while ``–'' indicates marginal or absent usage.
The database comprises 934 lexical items and 9,057 variants, covering 21 Spanish-speaking countries\footnote{Note that Puerto Rico, which is currently a self-governing commonwealth of the United States, was counted as a country for uniformity.}.

Each item includes an index\footnote{A first letter in the item index, either of  ``A'',  ``B'',  ``C'',  ``D'',  ``E'', ``F'', ``I'', represents when the original survey was conducted and does not hold linguistic relevance.}, a Spanish description\footnote{Example sentences included in some entries were excluded for uniformity.}, an English gloss, and a list of lexical variants.
Table~\ref{tab:A141} illustrates the structure of a representative item (A141), referring to a vehicle equivalent to English \textit{car}, for which VARILEX lists six variants (\textit{auto}, \textit{automóvil}, \textit{carro}, \textit{coche}, \textit{concho}, \textit{máquina}).
The number of variants per item varies substantially, ranging from 1 to 114, with a median of 8.
Moreover, the number of predominant variants differs across countries per item; while Spain predominantly uses only one variant (\textit{coche}) for item A141, Argentina exhibits multiple predominant forms (\textit{auto}, \textit{automóvil}, \textit{coche}).
These characteristics make VARILEX particularly suitable for evaluating fine-grained geographic lexical knowledge, as they reflect both cross-item and cross-country variation in lexical inventories.

\begin{table}[tb]
    \small
    \centering
    \begin{tabular}{lc}
        \hline
        Item index & A141 \\
        \hline
        \hline
        Description &
            \begin{tabular}{c}
                \textit{vehículo destinado} \\
                \textit{al transporte de personas} \\
                ``\textit{vehicle intended} \\
                  \textit{for the transport of people}''
            \end{tabular}
            \\
        \hline
        English equivalent & CAR\\
        \hline
        Variants &
            \begin{tabular}{c}
                \textit{auto}, \textit{automóvil}, \textit{carro}, \\
                \textit{coche}, \textit{concho}, \textit{máquina}
            \end{tabular}
            \\
        \hline    
      \end{tabular}
    \caption{Item A141 referring to \textit{car}.}
    \label{tab:A141}
\end{table}

\subsection{Question Formats}
\label{subsec:questions}
To probe lexical variation, we employ two complementary question–answering formats that simulate traditional dialectological surveys.
See Appendix~\ref{sec:appendix_template} for question templates.

\paragraph{Yes–No Question format (YNQF)}
The \emph{Yes–No Question Format} simulates a survey in which an informant from a given country is asked whether they typically use a specific lexical variant to refer to a certain concept.
The model is instructed to respond exclusively with \textit{Sí} ``Yes'' or \textit{No} ``No''.
The predominant variants for each country are labeled ``+'' in the dataset.
Each prompt specifies the country, the lexical variant, and the Spanish description of the item.
This format allows us to assess the model’s ability to recognize whether a given variant is predominant in a particular country.

\paragraph{Multiple–Choice Question Format (MCQF)}
The \emph{Multiple–Choice Question Format}~\cite{li-etal-2024-multiple} simulates a survey in which an informant is asked to select all lexical variants they would typically use to refer to a given concept.
The variants associated with an item are enumerated as options, and the model is instructed to return only the corresponding option numbers, separated by a forward slash ``/'' if multiple options are selected.
The set of gold variants for each country comprises those labeled ``+'' in the dataset.
This format captures the fact that multiple variants may be simultaneously predominant within a country and enables set-based comparison between gold annotations and model responses.

The two formats offer complementary perspectives on regional lexical knowledge.
YNQF evaluates each variant independently, facilitating analysis of acceptance and rejection errors.
However, responses may be affected by a tendency to answer affirmatively or negatively.
MCQF instead evaluates the set of variants selected for a given concept and country, accommodating the coexistence of multiple regional alternatives.
Comparing formats helps assess the robustness of geographic performance patterns to the elicitation design.

\subsection{Evaluation Metrics}
\label{subsec:metrics}
We evaluate the performance of the models separately for YNQF and MCQF using metrics tailored to the structure of each task.

\paragraph{F1 Score}
For YNQF, we measure performance by means of the conventional F1 score.
Higher F1 scores indicate better alignment between model responses and gold annotations.

\paragraph{Adjusted Jaccard Coefficient}
For MCQF, we measure performance using the adjusted Jaccard coefficient ($J_{\mathrm{adj}}$), a chance-corrected measure of overlap between two sets.
By correcting for chance agreement, $J_{\mathrm{adj}}$ enables meaningful comparison between items with differing numbers of variants and supports robust aggregation at the country level.
Country–level performance is computed as the average across questions.

Let \textit{A} denote the set of gold-standard predominant variants for a given item and country, and \textit{B} the set of variants predicted by the model.
The conventional Jaccard coefficient~\cite{real-vargas-1996-probabilistic} is defined as:
\begin{equation}
    J(A,B) = \frac{|A \cap B|}{|A \cup B|}
           = \frac{X}{s + t - X},
\end{equation}
where $|A| = s$, $|B| = t$, and $X = |A \cap B|$.
A larger value means better performance.
However, because overlap can arise spuriously when sets are large, we correct for chance agreement by defining:
\begin{equation}
    J_{\mathrm{adj}}
        = \frac{J - \mathbb{E}[J]}{1 - \mathbb{E}[J]},
\end{equation}
where $J$ is the conventional Jaccard coefficient, and $\mathbb{E}[J]$ is its expected value.

Assume the null hypothesis that $A$ and $B$ are formed by selecting $s$ and $t$ variants uniformly at random from the $N$ variants, independently of each other.
Under this model, the size of the intersection $X$ follows a hypergeometric distribution:
\begin{equation*}
    X \sim \mathrm{Hypergeometric}(N, t, s).
\end{equation*}

Thus,
\begin{equation*}
    \mathbb{E}[X] = \frac{s t}{N}.
\end{equation*}
This expected value quantifies the chance similarity arising from the random overlap between two randomly selected sets of sizes $s$ and $t$.

Because $|A \cup B| = s + t - X$, we obtain
\begin{equation*}
    \mathbb{E}[|A \cup B|]
      = s + t - \mathbb{E}[X]
      = s + t - \frac{s t}{N}.
    \end{equation*}

Therefore, a common first-order approximation to the expected Jaccard coefficient is
\begin{equation}
    \begin{aligned}
        \mathbb{E}[J]
          \approx
          \frac{\mathbb{E}[X]}
               {\mathbb{E}[|A \cup B|]}
          &=
          \frac{\displaystyle \frac{s t}{N}}
               {\displaystyle s + t - \frac{s t}{N}} \\
          &=
          \frac{s t}
               {N(s + t) - s t}.
    \end{aligned}
\end{equation}
The adjusted Jaccard coefficient $J_{\mathrm{adj}}$ provides a chance-corrected, normalized measure of similarity between two sets.

This transformation has the following properties:
\begin{itemize}
    \item $J_{\mathrm{adj}} = 1$ if and only if $A = B$.
    \item $J_{\mathrm{adj}} = 0$ when the observed similarity equals its expected value under the null model.
    \item $J_{\mathrm{adj}} < 0$ indicates less overlap than would be expected by chance.
\end{itemize}
We clipped negative values to zero, ensuring that the metric is bounded between 0 and 1.

\section{Experiments}
\label{sec:experiment}

\subsection{Experimental Setting}
\label{subsec:setting}

\paragraph{Prompts}
For YNQF, we selected 242 lexical items across the countries studied, ensuring that each country-specific sample included both predominant (positive) and marginal (negative) variants.
Thus, each country-specific sample comprised 484 questions, evenly divided into 242 positive and 242 negative questions, yielding a total of 10,164 questions across 21 countries.
For MCQF, we used 16,637 questions corresponding to lexical items for which the number of variants are between two and twenty.
To mitigate ordering effects, the variants associated with each item were randomly shuffled before enumeration in MCQF, following established findings on option-order sensitivity in multiple–choice evaluation~\cite{pezeshkpour-hruschka-2024-large}.
In both formats, questions were presented to the models in randomized order.
Only responses that strictly adhered to the prompt instructions (i.e., valid \textit{Sí/No} answers for YNQF and properly formatted numeric selections for MCQF) were retained for evaluation.

\paragraph{Baselines}
For YNQF, we use a uniform random baseline that independently predicts Yes or No with equal probability.
With balanced gold labels, this yields theoretical accuracy, precision, recall, and F1 scores of 0.500.
For MCQF, we set the baseline by selecting the first \textit{k} options (e.g., 1/2/3 for $k=3$) for a given item.
The value of \(k\) represents the expected number of predominant variants for an item.
It is calculated by multiplying the number of variants for that item by 0.286—the median proportion of predominant variants across items—and rounding to the nearest integer.

\paragraph{Models}
We evaluate ten models, comprising three proprietary and seven open-weight models.
Full model identifiers are provided in Table~\ref{tab:model-ids} in Appendix~\ref{appendix:model-ids}.
\begin{itemize}[leftmargin=*, nosep]
    \item \textbf{Proprietary:} Claude Haiku 4.5, Gemini 3.5 Flash Lite, and GPT-4o.
    \item \textbf{Open-weight:} Gemma 4 12B, Gemma 4 31B, Llama 3.1 8B, Ministral 3 8B, Mistral Small 24B, Qwen2.5 7B, and Qwen3 30B-A3B.
\end{itemize}
The proprietary models were accessed via the official API, and the open-weight models were locally downloaded from \texttt{Hugging Face}\footnote{\url{https://huggingface.co/}}.
The temperature was set to 0 for all models to minimize variability and promote deterministic responses.

\subsection{Results}
\label{subsec:results}

\subsubsection{YNQF}
\label{subsubsec:results_YN}
Table~\ref{tab:results_YN} summarizes overall performance on YNQF.
It shows substantial variation in YNQF performance despite nearly universal response validity.
GPT-4o achieved the highest accuracy (0.700), while Gemini 3.5 Flash Lite obtained the highest $F1$ (0.696).
Precision–recall differences reveal contrasting response tendencies: Claude Haiku 4.5 achieved the highest precision (0.864) but extremely low recall (0.015), indicating highly conservative positive predictions.
Gemma 4 12B and Mistral Small 24B exhibited similar patterns.
Conversely, Ministral 3 8B attained the highest recall (0.910) but relatively low precision (0.515), suggesting broad acceptance of candidate expressions.
Although all models numerically exceeded the baseline accuracy of 0.500, only six surpassed its $F1$ of 0.500, highlighting the importance of examining precision and recall alongside accuracy.

\begin{table*}[t]
  \centering
  \begingroup
  \small
  \setlength{\tabcolsep}{8pt}
  \renewcommand{\arraystretch}{1.1}
  \begin{tabular}{lccccc}
    \toprule
    Model & Valid ratio & Accuracy & Precision & Recall & $F1$ \\
    \midrule
    Baseline & 1.000 & 0.500 & 0.500 & 0.500 & 0.500 \\
    \hline
    Claude Haiku 4.5
      & 1.000 & 0.506 & \textbf{0.864} & 0.015 & 0.029 \\
    Gemini 3.5 Flash Lite
      & 1.000 & 0.667 & 0.640 & 0.763 & \textbf{0.696} \\
    GPT-4o
      & 1.000 & \textbf{0.700} & 0.718 & 0.659 & 0.687 \\
    Gemma 4 12B
      & 1.000 & 0.522 & 0.776 & 0.063 & 0.116 \\
    Gemma 4 31B
      & 1.000 & 0.656 & 0.671 & 0.612 & 0.640 \\
    Llama 3.1 8B
      & 1.000 & 0.573 & 0.569 & 0.605 & 0.587 \\
    Ministral 3 8B
      & 1.000 & 0.527 & 0.515 & \textbf{0.910} & 0.658 \\
    Mistral Small 24B
      & 0.996 & 0.517 & 0.832 & 0.043 & 0.082 \\
    Qwen2.5 7B
      & 1.000 & 0.542 & 0.618 & 0.219 & 0.324 \\
    Qwen3 30B-A3B
      & 1.000 & 0.591 & 0.571 & 0.729 & 0.641 \\
    \bottomrule
  \end{tabular}
  \endgroup
  \caption{Overall results for the Yes--No question format
    (YNQF). Valid ratio denotes the proportion of valid
    responses.}
  \label{tab:results_YN}
\end{table*}

\begin{figure*}[tb]
  \includegraphics[width=\textwidth]{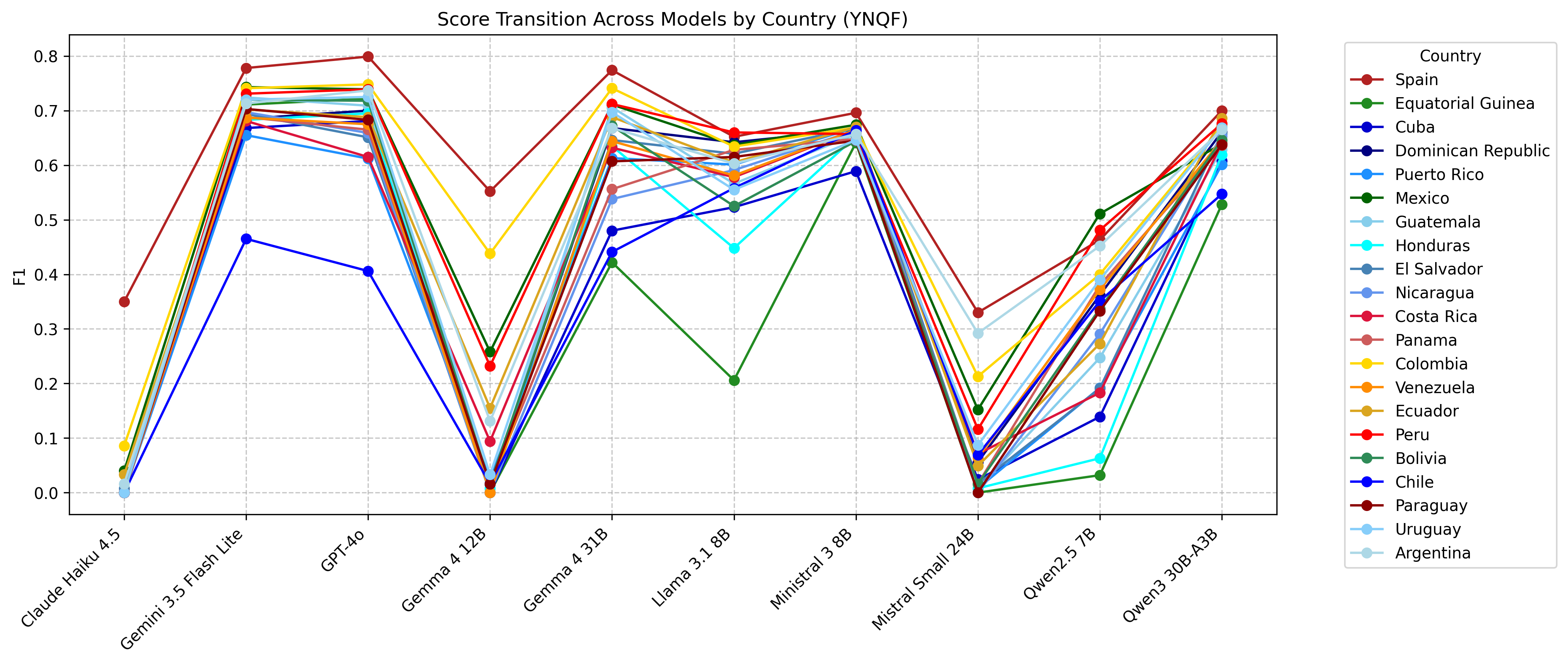}
  \caption{Score Transition Across Models by Country (YNQF).}
  \label{fig:scores_country_across_models_YN}
\end{figure*}

Figure~\ref{fig:scores_country_across_models_YN} illustrates score transition across models by country.
See Table~\ref{tab:scores_country_across_models_YN} in Appendix~\ref{sec:supplementary_tables} for the specific values.
Country-level performance varied substantially across models.
Spain received the highest score for eight of the ten models, including GPT-4o (0.799), Gemini 3.5 Flash Lite (0.778), and Gemma 4 31B (0.774).
The exceptions were Llama 3.1 8B, which performed best for Peru (0.660), and Qwen2.5 7B, which peaked for Mexico (0.511).
Averaging scores equally across models, Spain ranked highest, followed by Colombia and Mexico, while Equatorial Guinea and Chile ranked lowest.
However, these patterns were not uniform: Ministral 3 8B showed relatively little geographic variation (0.589–0.696), whereas Claude Haiku 4.5 scored zero for 15 countries.

\subsubsection{MCQF}
\label{subsubsec:results_MC}
Table~\ref{tab:results_MC} presents overall performance on MCQF.
All models exceeded the baseline adjusted Jaccard score of 0.137.
GPT-4o achieved the highest score (0.346), followed by Gemini 3.5 Flash Lite (0.334) and Gemma 4 31B (0.325), the strongest open-weight model.
Qwen2.5 7B obtained the lowest score (0.210), although it remained above the baseline.
Response validity was consistently high (0.986–1.000), suggesting that invalid outputs were relatively uncommon and that the principal performance differences concerned the selected lexical variants.
Overall performance did not significantly correlate with that of YNQF (Spearman's $\rho=0.285$ ($p=0.425$))\footnote{This study uses Spearman's rank correlation $\rho$ and sets the significance threshold at $p = 0.05$.}.
The number of variants per item correlates negatively with model performance, indicating that items with more variants tend to be more difficult for the models ($\rho$ ranges from $-0.473$ to $-0.545$ ($p < 0.001$)). 

\begin{table}[t]
  \centering
  \begingroup
  \small
  \setlength{\tabcolsep}{5pt}
  \renewcommand{\arraystretch}{1.1}
  \resizebox{\linewidth}{!}{%
    \begin{tabular}{lccc}
      \toprule
      Model & Valid ratio & $J$ & $J_{adj}$\\
      \midrule
      Baseline & 1.000 & 0.215 & 0.137 \\
      \hline
      Claude Haiku 4.5      & 1.000 & 0.393 & 0.259 \\
      Gemini 3.5 Flash Lite & 1.000 & 0.430 & 0.334 \\
      GPT-4o               & 1.000 & \textbf{0.467} & \textbf{0.346} \\
      Gemma 4 12B          & 1.000 & 0.410 & 0.269 \\
      Gemma 4 31B          & 1.000 & 0.445 & 0.325 \\
      Llama 3.1 8B         & 0.995 & 0.445 & 0.221 \\
      Ministral 3 8B       & 0.986 & 0.300 & 0.214 \\
      Mistral Small 24B    & 0.999 & 0.322 & 0.263 \\
      Qwen2.5 7B           & 0.996 & 0.332 & 0.210 \\
      Qwen3 30B-A3B        & 1.000 & 0.387 & 0.225 \\
      \bottomrule
    \end{tabular}%
  }
  \endgroup
  \caption{Overall results for the multiple-choice question format (MCQF). Valid ratio denotes the proportion of valid responses. $J$ represents the original Jaccard scores.
    }
  \label{tab:results_MC}
\end{table}

\begin{figure*}[tb]
  \includegraphics[width=2\columnwidth]{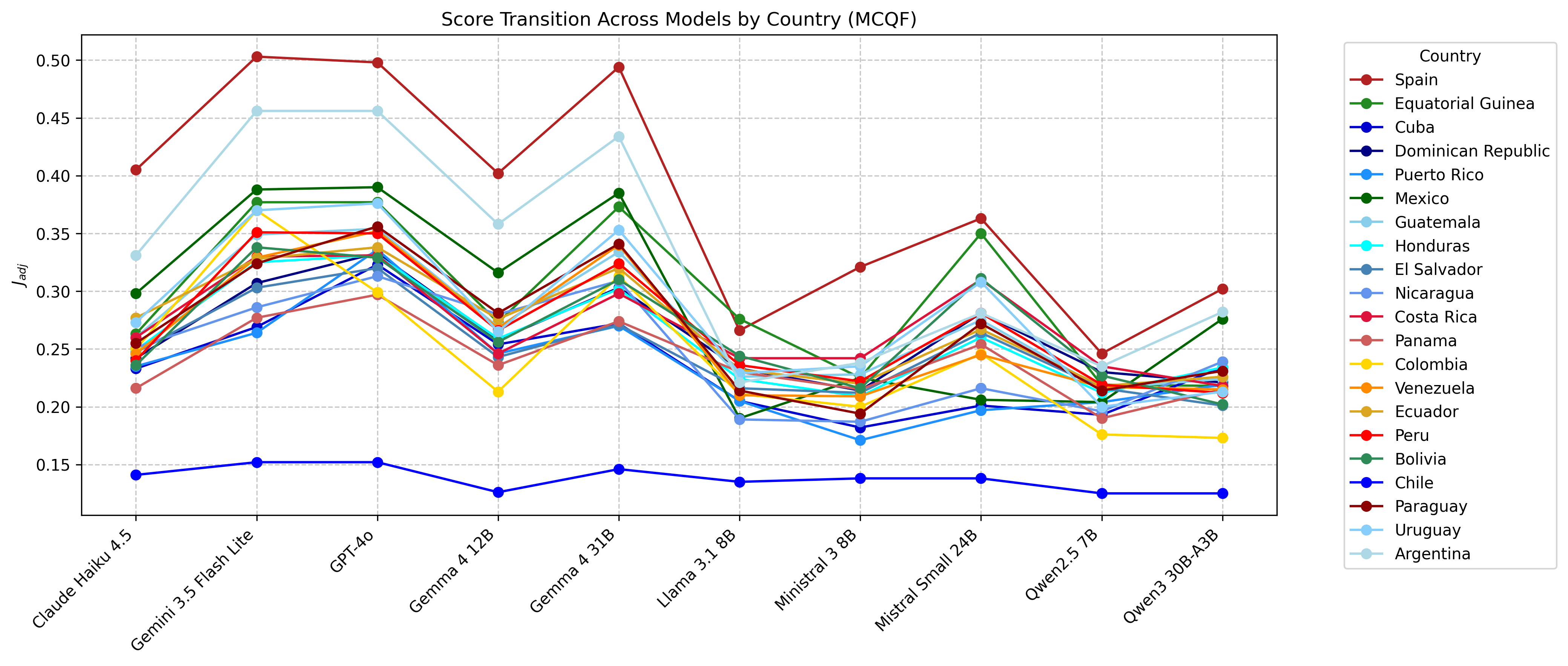}
  \caption{Score Transition Across Models by Country (MCQF).}
  \label{fig:scores_country_across_models_MC}
\end{figure*}

Figure~\ref{fig:scores_country_across_models_MC} illustrates score transition across models by country on MCQF.
See Table~\ref{tab:scores_country_across_models_MC} in Appendix~\ref{sec:supplementary_tables} for the specific values.
Geographic patterns were more consistent across models in MCQF.
Spain received the highest score for nine models, with Llama 3.1 8B instead performing best for Equatorial Guinea (0.276).
Chile received the lowest score for every model, ranging from 0.125 to 0.152.
Equal-weight averages across models placed Spain first (0.380), followed by Argentina (0.329) and Equatorial Guinea (0.295).
Equatorial Guinea therefore showed a particularly pronounced contrast between formats, ranking last on average in YNQF but third in MCQF.
These differences indicate that country rankings depend on elicitation format, although scores from the two formats are not directly comparable.

It is noteworthy that adjusting Jaccard substantially altered its association with the chance baseline (Table~\ref{tab:jaccard-correlation-comparison} in Appendix~\ref{sec:supplementary_tables}).
Across the ten models evaluated with the original and adjusted Jaccard, the number of significant correlations decreased from eight to three at the uncorrected 0.05 level.
Strong positive associations largely disappeared for GPT-4o, Claude Haiku 4.5, and Qwen2.5 7B.
However, adjustment did not uniformly eliminate the association: Llama 3.1 8B exhibited a pronounced reversal, Mistral Small 24B showed a stronger negative association, and Qwen3 30B-A3B retained a positive association.
These results suggest that adjustment reduces alignment with chance performance for several models while also changing the interpretation of geographic disparities.

\section{Discussion}
\label{sec:discussion}

\subsection{Inter-Model Correlations}
\label{sec:intermodel-correlation}
To examine whether models exhibited similar geographic performance patterns, we computed Spearman's rank correlation coefficient between their country-level scores.
Figure~\ref{fig:inter-model correlation} in Appendix~\ref{sec:supplementary_figures} illustrates the inter-model correlations.

\paragraph{Agreement within formats}
Correlations were predominantly positive, indicating that many models shared geographic performance tendencies.
Within YNQF, Gemini 3.5 Flash Lite and GPT-4o showed strong agreement ($\rho=0.87$), as did Gemini and Gemma 4 31B ($\rho=0.76$).
However, this agreement was not uniform: GPT-4o and Ministral 3 8B exhibited little correspondence ($\rho=0.04$).
Within MCQF, the strongest correlation occurred between Llama 3.1 8B and Mistral Small 24B ($\rho=0.91$).
Other pairs showed weak or slightly negative correlations, indicating that shared geographic patterns did not extend equally to all models.

\paragraph{Agreement across formats}
Within-model correlations between YNQF and MCQF were generally weaker.
Gemini showed the strongest cross-format agreement ($\rho=0.83$), followed by GPT-4o ($\rho=0.59$), Claude Haiku 4.5 ($\rho=0.57$), and Gemma 4 31B ($\rho=0.50$).
The remaining six models exhibited correlations below 0.20, including Llama 3.1 8B, Qwen2.5 7B, and Qwen3 30B-A3B, each with $\rho=0.04$.
Thus, for most models, countries that ranked relatively highly in one format did not consistently rank highly in the other.

\paragraph{Implications}
The combination of strong correlations among several models within formats and weak cross-format correlations for most individual models suggests that evaluation design shapes the geographic patterns observed.
YNQF elicits judgments about individual variants, whereas MCQF requires selection among simultaneously presented alternatives; these conditions may affect response tendencies and the use of regional lexical knowledge.
Nevertheless, the observed differences cannot be attributed to elicitation format alone, because scoring procedures and item composition may also contribute.
Moreover, high inter-model correlation does not establish high performance or shared underlying mechanisms: models can produce similar country rankings while differing substantially in overall accuracy.
These results therefore support evaluating geographic disparities through complementary formats and testing cross-format consistency on matched lexical items.

\subsection{Regression Analysis of Country-Level Performance}
\label{sec:regression}

\subsubsection{Methods}
To investigate the factors associated with geographic differences in performance, we fitted an ordinary least squares regression with an intercept for each model and question format.
The dependent variable was the country-level performance score, and the predictors were digital resource volume and lexical distance from Spain described below.
To facilitate coefficient comparisons within each regression, both predictors were standardized across countries to zero mean and unit variance.

\paragraph{Lexical distance from Spain}
We measured countries' lexical distance with respect to the best-performing Spain on both question formats.
The distance was computed using Jaccard distance, based on the original positive/negative annotations for YNQF and synonym sets for MCQF (Tables~\ref{tab:jaccard-distance-spain-YN}–\ref{tab:jaccard-distance-spain-MC} in Appendix~\ref{sec:supplementary_tables}).

\paragraph{Digital Resources}
As the actual data volume used to train LLMs is not made public, following~\citet{mayorrocher-et-al-2025-igual}, we used as a proxy the estimated volume of digital resources derived from CEREAL (Corpus del Español REAL)~\cite{espanabonet-barroncedeno-2024-elote}\footnote{\url{https://zenodo.org/records/14771240}}.
The volume was measured in terms of tokens (Tables~\ref{tab:country-data-volume} in Appendix~\ref{sec:supplementary_tables}) and then log10-transformed to reduce skewness.

\subsubsection{Relative associations of the predictors}
\paragraph{YNQF}
In YNQF, digital resource volume generally exhibited larger absolute coefficients than lexical distance (Figure~\ref{fig:regression_coefficients_YN} in Appendix~\ref{sec:supplementary_figures}).
YNQF presents each expression separately, requiring the model to decide whether it is used in the specified country without access to competing alternatives.
Greater exposure to country-associated vocabulary could make these judgments easier, providing a possible explanation for the stronger association with digital resource volume.

\paragraph{MCQF}
In MCQF, lexical distance generally exhibited the larger coefficients (Figure~\ref{fig:regression_coefficients_MC} in Appendix~\ref{sec:supplementary_figures}).
MCQF presents alternative expressions together, providing cues that may facilitate recognition while requiring models to discriminate among their regional distributions.
If models rely on lexical associations that align more closely with Spain, greater lexical distance from Spain could make this selection more difficult.
This explanation would be supported by negative distance coefficients.

\subsubsection{Explained variation across models}
In YNQF, models with lower overall performance tended to exhibit higher adjusted $R^2$ (Figure~\ref{fig:scatter_performance_adjR2_YN} in Appendix~\ref{sec:supplementary_figures}).
Thus, the two predictors jointly accounted for a larger proportion of cross-country performance variation in these models.
This pattern should not be interpreted as evidence that weaker models depend more heavily on resource volume alone: adjusted $R^2$ summarizes the explanatory contribution of the full regression, including lexical distance.
% Nor does it directly establish weaker generalization.
No comparable relationship between overall performance and adjusted $R^2$ was observed in MCQF, suggesting that the alignment between performance disparities and the selected predictors differed across elicitation formats (Figure~\ref{fig:scatter_performance_adjR2_MC} in Appendix~\ref{sec:supplementary_figures}).

\subsection{Relationship to Country-Level Performance}
\paragraph{Equatorial Guinea’s contrasting performance}
The combination of data paucity and lexical affinity with Peninsular Spanish may explain Equatorial Guinea’s contrasting performance across formats, ranking last on average in YNQF but third in MCQF: the smallest digital resources may hinder isolated usage judgments in YNQF, whereas vocabulary shared with Spain (the third closest) may facilitate selection among alternatives in MCQF.
This interpretation is consistent with the regression results, which show stronger associations with resource volume in YNQF and with lexical distance in MCQF.
Nevertheless, the contrast does not establish that models possess detailed knowledge of Equatoguinean usage: relatively strong MCQF performance could partly reflect reliance on shared vocabulary rather than country-specific distinctions.

\paragraph{Chile's low performance}
Chile combines relatively abundant digital resources with substantial lexical distance from Spain.
Its corpus size ranks fourth, whereas its lexical distance from Spain is the fourth- and third-largest in YNQF and MCQF, respectively.
Its low performance therefore cannot be readily attributed to low aggregate data availability.
Instead, this pattern is consistent with the regression results suggesting that MCQF performance is more closely associated with lexical distance than with resource volume: extensive digital representation may not ensure accurate recognition of vocabulary specific to Chile.
This finding is noteworthy given the well-documented lexical distinctiveness of Chilean Spanish~\cite{sanmartin-nunez-2023-espanol}, and implies a disconnect between traditional descriptions and the lexical patterns represented in web data (e.g., spoken vs. written language).

\section{Conclusions}
\label{sec:conclusions}
We evaluated ten LLMs’ geographic lexical knowledge across 21 Spanish-speaking countries using Yes–No and multiple-choice questions.
Performance varied across countries and question formats, with Spain generally achieving the highest scores.
Regression analyses associated YNQF performance more strongly with digital resource availability and MCQF performance with lexical distance from Spain.
Adjusted Jaccard scoring reduced, but did not eliminate, associations with baseline difficulty.
These findings highlight the importance of elicitation format and evaluation metrics when interpreting geographic disparities.
Future work should examine whether these patterns extend to spontaneous language generation and practical dialect-sensitive applications.

\section*{Limitations}
\label{sec:limitations}
This study has several limitations that point to directions for future research.
First, treating LLMs as virtual informants implicitly assumes that their responses approximate those of an average speaker of each dialect.
In doing so, we abstract away from sociolinguistic factors such as age, gender, educational background, and social context, all of which are known to influence linguistic choice~\cite{holmes-2008-introduction}.
Explicitly encoding such variables in prompts may affect variant selection and warrant systematic investigation.

Second, for analytical simplicity, we treat countries as the minimal units of geographic variation and assume internal homogeneity within national boundaries.
This abstraction overlooks well-documented intra-country variation, including regional, urban–rural, and social distinctions~\cite{gonçalves-sanchez-2014-crowdsourcing}.
Future work could refine this approach by incorporating subnational dialectal divisions or by modeling graded variation rather than categorical country–level usage.

Third, the evaluation methodology warrants further refinement.
Despite expert curation, annotation artifacts and VARILEX’s binary, country-level labels may contribute to observed disparities.
Adjusted Jaccard reduced associations with baseline difficulty, but residual and reversed correlations highlight the need to further assess and improve the correction.

Furthermore, our regression analyses are exploratory and rely on only 21 country-level observations, limiting statistical power and potentially making estimates sensitive to individual countries.
CEREAL token counts serve as a proxy for digital resource availability and may not reflect the models’ actual training data.
Spain was chosen as a descriptive reference because of its generally high performance, rather than as a linguistic norm.
The estimated associations should therefore not be interpreted causally, and nonsignificant results do not establish the absence of a relationship.

In addition to the variant recognition task, we should extend this evaluation to region-conditioned generation, examining whether knowledge of regional usage translates into productive lexical choices.
Comparing recognition and generation performance will clarify the relationship between explicit lexical judgments and context-appropriate production, providing a more comprehensive view of how linguistic diversity is represented in model outputs~\cite{dhamala-etal-2021-bold}.
Such an approach may also support the development of synthetic data for underrepresented varieties~\cite{munozbasols-et-al-2024-sesgo} and facilitate inter-dialect adaptation in practical applications.

Finally, other dimensions of linguistic variation, such as morphosyntactic~\cite{takagaki-etal-varigrama} and discourse-level phenomena, are equally central to dialectal differentiation and should be examined using comparable evaluation frameworks.

\section*{Ethical Considerations}
\label{sec:ethical}
This study involves no new human participants, personal data, or human annotation.
All analyses are based on publicly available linguistic resources (i.e., VARILEX) and responses generated using official APIs for proprietary models and locally downloaded open-weight models.

The work evaluates how geographic lexical variation is represented in language models and does not involve profiling, classification, or inference about individual speakers.
Countries and dialectal areas are used as analytical abstractions for evaluation purposes and should not be interpreted as homogeneous linguistic entities.
The reported results do not constitute normative judgments about varieties or their speakers and should not be construed as assessments of linguistic quality or legitimacy.

Our analysis is diagnostic in nature and aims to identify potential sources of performance divergence arising from data distributions and modeling choices.
Because the evaluated models are proprietary and their training data are not transparent, our conclusions are limited to observed model behavior and do not make claims about internal representations.

We used AI-assisted tools to improve grammar, clarity, and style; paraphrase and polish the original content; assist with literature searches; and brainstorm ideas and generate draft text.
All AI-assisted content was critically reviewed and verified by the authors, who take full responsibility for the final manuscript.

\section*{Acknowledgments}
This work was supported by JSPS KAKENHI Grant Numbers JP23K12152 and JP26K00004.

% \section{Bib\TeX{} Files}
% \label{sec:bibtex}

% Unicode cannot be used in Bib\TeX{} entries, and some ways of typing special characters can disrupt Bib\TeX's alphabetization. The recommended way of typing special characters is shown in Table~\ref{tab:accents}.

% Please ensure that Bib\TeX{} records contain DOIs or URLs when possible, and for all the ACL materials that you reference.
% Use the \verb|doi| field for DOIs and the \verb|url| field for URLs.
% If a Bib\TeX{} entry has a URL or DOI field, the paper title in the references section will appear as a hyperlink to the paper, using the hyperref \LaTeX{} package.


\bibliography{export.bib}

\appendix
\section{Question Templates}
\label{sec:appendix_template}
Each question specifies three components in brackets: \textit{country}, \textit{variant}, and \textit{description}.

\subsection{Yes–No Question Format (YNQF)}
\label{subsec:appendix_template_YNQF}
\textit{
    Responda a la siguiente pregunta. \\
    No tenga en cuenta las preguntas anteriores, si las hay. \\
    Usted es de [\textbf{country}], ¿verdad? \\
    ¿Suele utilizar la expresión «\textbf{[variant]}» para referirse a «\textbf{[description]}»? \\
    Responda únicamente con «Sí» o «No». \\
    No se requiere aclaración alguna.
    } \\
\\
\textit{
    ``Answer the following question. \\
    Do not take the previous questions into account, if any. \\
    You are from \textbf{[country]}, aren't you? \\
    Do you usually use the term `\textbf{[variant]}' to refer to `\textbf{[description]}'? \\
    Answer only with `Yes' or `No'. \\
    No clarification is required.''
    }

\subsection{Multiple–Choice Question Format (MCQF)}
\label{subsec:appendix_template_MCQF}
\textit{
    Responda a la siguiente pregunta. \\
    No tenga en cuenta las preguntas anteriores, si las hay. \\
    Usted es de [\textbf{country}], ¿verdad? \\
    ¿Qué expresión(es) suele usar para referirse a «\textbf{[description]}»?
    Las opciones son: \\
    1 \textbf{[variant 1]} \\
    2 \textbf{[variant 2]} \\
    3 \textbf{[variant 3]} \\
    \ldots \\
    Conteste solo con el número correspondiente a la opción. \\
    Puede elegir más de una opción; en ese caso, los números deberán ir separados por el signo «/» en orden ascendente. \\
    No se requiere aclaración alguna.
    } \\
\\
\textit{
    ``Answer the following question. \\
    Do not take the previous questions into account, if any. \\
    You are from \textbf{[country]}, aren't you? \\
    Which expression(s) do you usually use to refer to `\textbf{[description]}'?
    The options are: \\
    1 \textbf{[variant 1]} \\
    2 \textbf{[variant 2]} \\
    3 \textbf{[variant 3]} \\
    \ldots \\
    Answer only with the number(s) corresponding to the selected option(s). \\
    You may choose more than one option; in that case, the numbers should be separated by the `/' sign in ascending order. \\
    No clarification is required.''
    }

\section{Model Identifiers}
\label{appendix:model-ids}

\begin{table}[htb]
  \centering
  \begingroup
  % \small
  \setlength{\tabcolsep}{6pt}
  \renewcommand{\arraystretch}{1.1}
  \resizebox{\columnwidth}{!}{%
    \begin{tabular}{ll}
      \toprule
      Model & Full identifier \\
      \midrule
      \multicolumn{2}{l}{\textbf{Proprietary models}} \\
      \addlinespace
      Claude Haiku 4.5
        & \texttt{claude-haiku-4-5-20251001} \\
      Gemini 3.5 Flash Lite
        & \texttt{gemini-3.5-flash-lite} \\
      GPT-4o
        & \texttt{gpt-4o} \\
      \midrule
      \multicolumn{2}{l}{\textbf{Open-weight models}} \\
      \addlinespace
      Gemma 4 12B
        & \texttt{google/gemma-4-12B} \\
      Gemma 4 31B
        & \texttt{google/gemma-4-31B} \\
      Llama 3.1 8B
        & \texttt{meta-llama/Meta-Llama-3.1-8B-Instruct} \\
      Ministral 3 8B
        & \texttt{mistralai/Ministral-3-8B-Instruct-2512} \\
      Mistral Small 24B
        & \texttt{mistralai/Mistral-Small-24B-Instruct-2501} \\
      Qwen2.5 7B
        & \texttt{Qwen/Qwen2.5-7B-Instruct} \\
      Qwen3 30B-A3B
        & \texttt{Qwen/Qwen3-30B-A3B-Instruct-2507} \\
      \bottomrule
    \end{tabular}%
  }
  \endgroup
  \caption{Abbreviated model names and full identifiers.}
  \label{tab:model-ids}
\end{table}

\onecolumn
\section{Supplementary Tables}
\label{sec:supplementary_tables}

\begin{table*}[htb]
  \centering
  \begingroup
  \small
  \setlength{\tabcolsep}{4pt}
  \renewcommand{\arraystretch}{1.1}
  \resizebox{\textwidth}{!}{%
    \begin{tabular}{c*{10}{r}}
      \toprule
      Country
      & \shortstack{Claude\\Haiku 4.5}
      & \shortstack{Gemini 3.5\\Flash Lite}
      & GPT-4o
      & \shortstack{Gemma 4\\12B}
      & \shortstack{Gemma 4\\31B}
      & \shortstack{Llama 3.1\\8B}
      & \shortstack{Ministral 3\\8B}
      & \shortstack{Mistral Small\\24B}
      & \shortstack{Qwen2.5\\7B}
      & \shortstack{Qwen3\\30B-A3B} \\
      \midrule
      Spain & \textbf{0.350} & \textbf{0.778} & \textbf{0.799} & \textbf{0.552} & \textbf{0.774} & 0.652 & \textbf{0.696} & \textbf{0.330} & 0.463 & \textbf{0.700} \\
      Equatorial Guinea & 0.000 & 0.711 & 0.722 & 0.000 & 0.422 & 0.206 & 0.641 & 0.000 & 0.032 & 0.528 \\
      Cuba & 0.000 & 0.668 & 0.681 & 0.000 & 0.480 & 0.523 & 0.589 & 0.024 & 0.139 & 0.608 \\
      Dominican Republic & 0.000 & 0.685 & 0.700 & 0.016 & 0.668 & 0.642 & 0.665 & 0.056 & 0.362 & 0.661 \\
      Puerto Rico & 0.000 & 0.655 & 0.612 & 0.024 & 0.613 & 0.601 & 0.669 & 0.008 & 0.192 & 0.601 \\
      Mexico & 0.040 & 0.743 & 0.739 & 0.258 & 0.711 & 0.637 & 0.674 & 0.152 & \textbf{0.511} & 0.639 \\
      Guatemala & 0.000 & 0.724 & 0.709 & 0.008 & 0.706 & 0.566 & 0.657 & 0.008 & 0.247 & 0.629 \\
      Honduras & 0.000 & 0.683 & 0.695 & 0.008 & 0.635 & 0.448 & 0.653 & 0.008 & 0.063 & 0.619 \\
      El Salvador & 0.008 & 0.694 & 0.651 & 0.000 & 0.646 & 0.621 & 0.668 & 0.016 & 0.191 & 0.658 \\
      Nicaragua & 0.000 & 0.697 & 0.659 & 0.000 & 0.538 & 0.591 & 0.664 & 0.000 & 0.291 & 0.653 \\
      Costa Rica & 0.000 & 0.681 & 0.615 & 0.094 & 0.632 & 0.578 & 0.665 & 0.071 & 0.183 & 0.640 \\
      Panama & 0.000 & 0.688 & 0.665 & 0.016 & 0.556 & 0.628 & 0.648 & 0.016 & 0.380 & 0.636 \\
      Colombia & 0.086 & 0.741 & 0.748 & 0.438 & 0.741 & 0.634 & 0.669 & 0.213 & 0.400 & 0.671 \\
      Venezuela & 0.000 & 0.687 & 0.675 & 0.000 & 0.644 & 0.581 & 0.665 & 0.064 & 0.372 & 0.646 \\
      Ecuador & 0.033 & 0.702 & 0.688 & 0.155 & 0.688 & 0.605 & 0.670 & 0.049 & 0.273 & 0.686 \\
      Peru & 0.000 & 0.731 & 0.739 & 0.232 & 0.712 & \textbf{0.660} & 0.657 & 0.116 & 0.481 & 0.676 \\
      Bolivia & 0.000 & 0.720 & 0.718 & 0.016 & 0.672 & 0.525 & 0.644 & 0.016 & 0.337 & 0.646 \\
      Chile & 0.000 & 0.465 & 0.406 & 0.016 & 0.441 & 0.558 & 0.664 & 0.069 & 0.351 & 0.547 \\
      Paraguay & 0.000 & 0.703 & 0.683 & 0.016 & 0.607 & 0.615 & 0.645 & 0.000 & 0.333 & 0.638 \\
      Uruguay & 0.000 & 0.720 & 0.725 & 0.033 & 0.697 & 0.555 & 0.646 & 0.087 & 0.390 & 0.669 \\
      Argentina & 0.016 & 0.713 & 0.737 & 0.131 & 0.668 & 0.603 & 0.657 & 0.292 & 0.452 & 0.665 \\
      \bottomrule
    \end{tabular}%
  }
  \endgroup
  \caption{Country-level $F1$ scores in the Yes--No question
    format (YNQF).}
  \label{tab:scores_country_across_models_YN}
\end{table*}

\begin{table*}[htb]
  \centering
  \begingroup
  \small
  \setlength{\tabcolsep}{4pt}
  \renewcommand{\arraystretch}{1.1}
  \resizebox{\textwidth}{!}{%
    \begin{tabular}{c*{10}{r}}
      \toprule
      Country
      & \shortstack{Claude\\Haiku 4.5}
      & \shortstack{Gemini 3.5\\Flash Lite}
      & GPT-4o
      & \shortstack{Gemma 4\\12B}
      & \shortstack{Gemma 4\\31B}
      & \shortstack{Llama 3.1\\8B}
      & \shortstack{Ministral 3\\8B}
      & \shortstack{Mistral Small\\24B}
      & \shortstack{Qwen2.5\\7B}
      & \shortstack{Qwen3\\30B-A3B} \\
      \midrule
      Spain & \textbf{0.405} & \textbf{0.503} & \textbf{0.498} & \textbf{0.402} & \textbf{0.494} & 0.266
      & \textbf{0.321} & \textbf{0.363} & \textbf{0.246} & \textbf{0.302} \\
      Equatorial Guinea & 0.263 & 0.377 & 0.377 & 0.273 & 0.373 & \textbf{0.276} & 0.225 & 0.350 & 0.219 & 0.218 \\
      Cuba & 0.233 & 0.269 & 0.323 & 0.254 & 0.272 & 0.205 & 0.182 & 0.201 & 0.193 & 0.234 \\
      Dominican Republic & 0.242 & 0.307 & 0.333 & 0.258 & 0.303 & 0.233 & 0.214 & 0.281 & 0.230 & 0.221 \\
      Puerto Rico & 0.235 & 0.264 & 0.336 & 0.246 & 0.270 & 0.205 & 0.171 & 0.197 & 0.204 & 0.219 \\
      Mexico & 0.298 & 0.388 & 0.390 & 0.316 & 0.385 & 0.190 & 0.225 & 0.206 & 0.204 & 0.276 \\
      Guatemala & 0.260 & 0.349 & 0.354 & 0.269 & 0.334 & 0.226 & 0.228 & 0.275 & 0.216 & 0.224 \\
      Honduras & 0.249 & 0.325 & 0.331 & 0.259 & 0.302 & 0.224 & 0.209 & 0.260 & 0.212 & 0.234 \\
      El Salvador & 0.247 & 0.303 & 0.320 & 0.243 & 0.271 & 0.216 & 0.212 & 0.264 & 0.216 & 0.201 \\
      Nicaragua & 0.251 & 0.286 & 0.313 & 0.280 & 0.309 & 0.189 & 0.187 & 0.216 & 0.196 & 0.239 \\
      Costa Rica & 0.260 & 0.330 & 0.331 & 0.246 & 0.298 & 0.242 & 0.242 & 0.310 & 0.235 & 0.218 \\
      Panama & 0.216 & 0.277 & 0.297 & 0.236 & 0.274 & 0.230 & 0.215 & 0.254 & 0.190 & 0.215 \\
      Colombia & 0.250 & 0.370 & 0.299 & 0.213 & 0.313 & 0.211 & 0.200 & 0.246 & 0.176 & 0.173 \\
      Venezuela & 0.246 & 0.329 & 0.352 & 0.268 & 0.340 & 0.210 & 0.209 & 0.245 & 0.217 & 0.215 \\
      Ecuador & 0.277 & 0.328 & 0.338 & 0.276 & 0.320 & 0.232 & 0.220 & 0.267 & 0.218 & 0.226 \\
      Peru & 0.240 & 0.351 & 0.350 & 0.266 & 0.324 & 0.236 & 0.222 & 0.281 & 0.219 & 0.212 \\
      Bolivia & 0.236 & 0.338 & 0.329 & 0.256 & 0.310 & 0.244 & 0.216 & 0.311 & 0.227 & 0.202 \\
      Chile & 0.141 & 0.152 & 0.152 & 0.126 & 0.146 & 0.135 & 0.138 & 0.138 & 0.125 & 0.125 \\
      Paraguay & 0.255 & 0.324 & 0.356 & 0.281 & 0.341 & 0.214 & 0.194 & 0.272 & 0.214 & 0.231 \\
      Uruguay & 0.273 & 0.370 & 0.376 & 0.265 & 0.353 & 0.229 & 0.235 & 0.308 & 0.200 & 0.213 \\
      Argentina & 0.331 & 0.456 & 0.456 & 0.358 & 0.434 & 0.221 & 0.238 & 0.281 & 0.235 & 0.282 \\
      \bottomrule
    \end{tabular}%
  }
  \endgroup
  \caption{Country-level $J_{adj}$ scores in the multiple-choice question
    format (MCQF).}
  \label{tab:scores_country_across_models_MC}
\end{table*}

% \twocolumn
\begin{table*}[htb]
  \centering
  \small
  \setlength{\tabcolsep}{8pt}
  \begin{tabular}{lrrrr}
    \toprule
    & \multicolumn{2}{c}{Original Jaccard}
    & \multicolumn{2}{c}{Adjusted Jaccard $J_{adj}$} \\
    \cmidrule(lr){2-3}\cmidrule(lr){4-5}
    Model & $\rho$ & $p$ & $\rho$ & $p$ \\
    \midrule
    GPT-4o
      & 0.656 & 0.001 & 0.077 & 0.739 \\
    Gemini 3.5 Flash Lite
      & 0.064 & 0.784 & -0.279 & 0.220 \\
    Claude Haiku 4.5
      & 0.787 & $<0.001$ & -0.086 & 0.710 \\
    Llama 3.1 8B
      & 0.640 & 0.002 & -0.679 & 0.001 \\
    Gemma 4 12B
      & 0.809 & $<0.001$ & 0.261 & 0.253 \\
    Gemma 4 31B
      & 0.448 & 0.042 & -0.098 & 0.674 \\
    Ministral 3 8B
      & 0.593 & 0.005 & -0.431 & 0.051 \\
    Mistral Small 24B
      & -0.181 & 0.431 & -0.580 & 0.006 \\
    Qwen2.5 7B
      & 0.832 & $<0.001$ & -0.149 & 0.518 \\
    Qwen3 30B-A3B
      & 0.829 & $<0.001$ & 0.462 & 0.035 \\
    \bottomrule
  \end{tabular}
  \caption{Spearman’s rank correlations $\rho$ between the performance and baseline under original and adjusted Jaccard scoring for MCQF. $p$-values are uncorrected for multiple comparisons.}
  \label{tab:jaccard-correlation-comparison}
\end{table*}

\begin{table}[htb]
  \centering
  \small
  \begin{tabular}{lr}
    \toprule
    Country & Jaccard distance from Spain \\
    \midrule
    España               & 0.000 \\
    Guinea Ecuatorial    & 0.668 \\
    Cuba                 & 0.584 \\
    República Dominicana & 0.697 \\
    Puerto Rico          & 0.627 \\
    México               & 0.640 \\
    Guatemala            & 0.745 \\
    Honduras             & 0.692 \\
    El Salvador          & 0.802 \\
    Nicaragua            & 0.640 \\
    Costa Rica           & 0.853 \\
    Panamá               & 0.734 \\
    Colombia             & 0.889 \\
    Venezuela            & 0.696 \\
    Ecuador              & 0.722 \\
    Perú                 & 0.769 \\
    Bolivia              & 0.756 \\
    Chile                & 0.795 \\
    Paraguay             & 0.649 \\
    Uruguay              & 0.767 \\
    Argentina            & 0.625 \\
    \bottomrule
  \end{tabular}
  \caption{Jaccard distances from Spain (YNQF).}
  \label{tab:jaccard-distance-spain-YN}
\end{table}

\begin{table}[htb]
  \centering
  \small
  \begin{tabular}{lr}
    \toprule
    Country & Jaccard distance from Spain \\
    \midrule
    España               & 0.000 \\
    Cuba                 & 0.545 \\
    Guinea Ecuatorial    & 0.554 \\
    Argentina            & 0.579 \\
    Puerto Rico          & 0.582 \\
    México               & 0.600 \\
    Paraguay             & 0.609 \\
    Nicaragua            & 0.609 \\
    Honduras             & 0.624 \\
    República Dominicana & 0.624 \\
    Venezuela            & 0.632 \\
    Ecuador              & 0.654 \\
    Guatemala            & 0.667 \\
    Bolivia              & 0.672 \\
    Perú                 & 0.678 \\
    Uruguay              & 0.681 \\
    Panamá               & 0.693 \\
    El Salvador          & 0.725 \\
    Chile                & 0.773 \\
    Costa Rica           & 0.792 \\
    Colombia             & 0.851 \\
    \bottomrule
  \end{tabular}
  \caption{Jaccard distances from Spain (MCQF).}
  \label{tab:jaccard-distance-spain-MC}
\end{table}

\begin{table}[t]
  \centering
  \small
  \begin{tabular}{lr}
    \toprule
    Country & Number of tokens \\
    \midrule
    Spain                & 48,821,356,834 \\
    Argentina            & 12,801,219,868 \\
    Mexico               & 10,829,453,655 \\
    Chile                &  6,589,894,993 \\
    Colombia             &  4,873,779,254 \\
    Peru                 &  2,868,606,294 \\
    Uruguay              &  1,212,536,074 \\
    Venezuela            &  1,123,220,785 \\
    Cuba                 &  1,107,823,343 \\
    Ecuador              &    835,942,678 \\
    Dominican Republic   &    626,587,024 \\
    Bolivia              &    411,271,023 \\
    Costa Rica           &    368,633,987 \\
    Paraguay             &    361,641,945 \\
    Honduras             &    300,205,673 \\
    Guatemala            &    278,048,894 \\
    Panama               &    255,371,136 \\
    Nicaragua            &    236,102,524 \\
    El Salvador          &    206,402,603 \\
    Puerto Rico          &     73,830,066 \\
    Equatorial Guinea    &     20,411,458 \\
    \bottomrule
  \end{tabular}
  \caption{Digital resource volume by country, measured
    in tokens and sorted in descending order.}
  \label{tab:country-data-volume}
\end{table}

\onecolumn
\section{Supplementary Figures}
\label{sec:supplementary_figures}

\begin{figure*}[htb]
  \includegraphics[width=\textwidth]{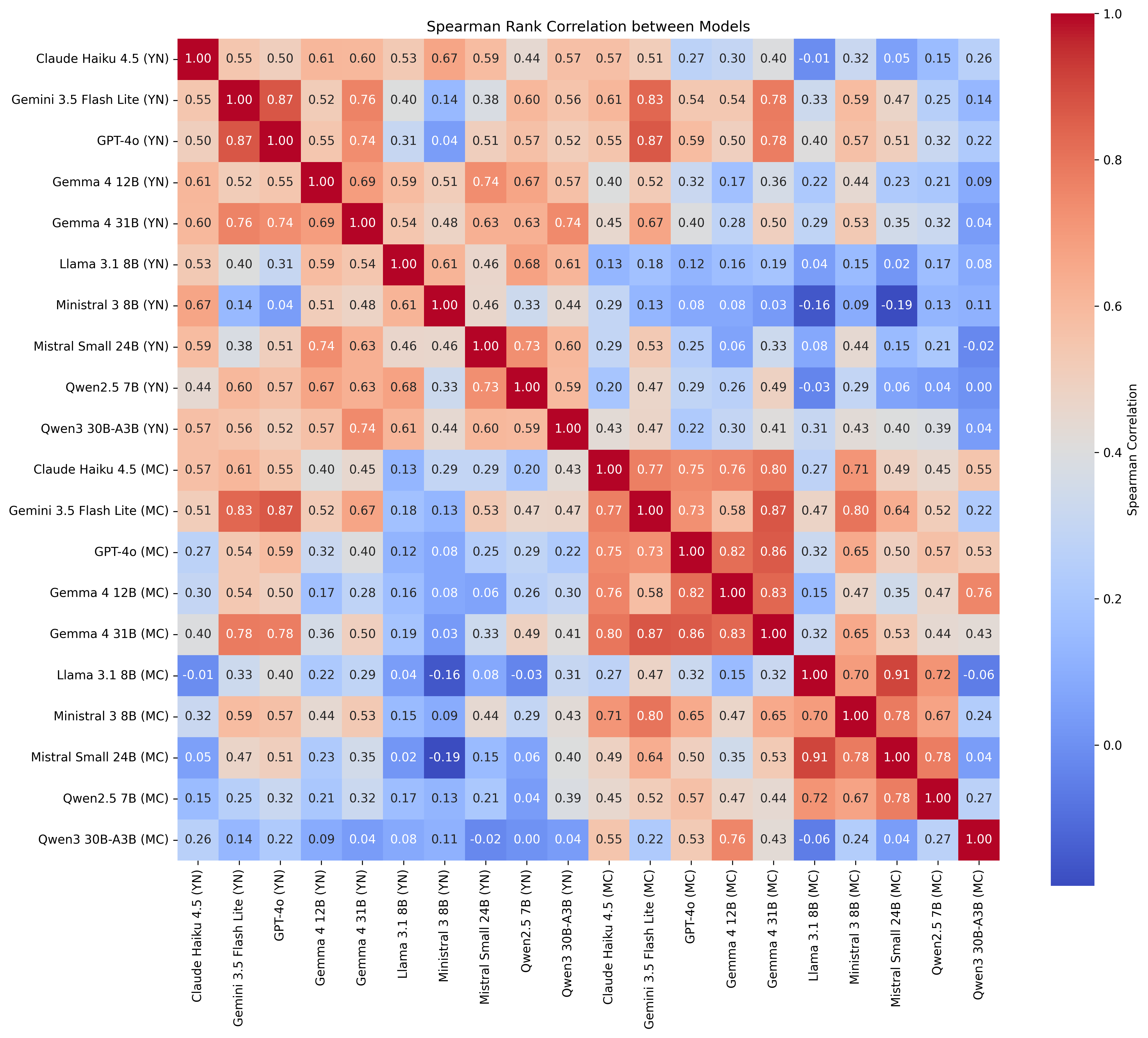}
  \caption{Inter-model correlation computed from their country-level performance scores. YN and MC represents YNQF and MCQF, respectively.}
  \label{fig:inter-model correlation}
\end{figure*}

\begin{figure*}[htb]
  \includegraphics[width=\textwidth]{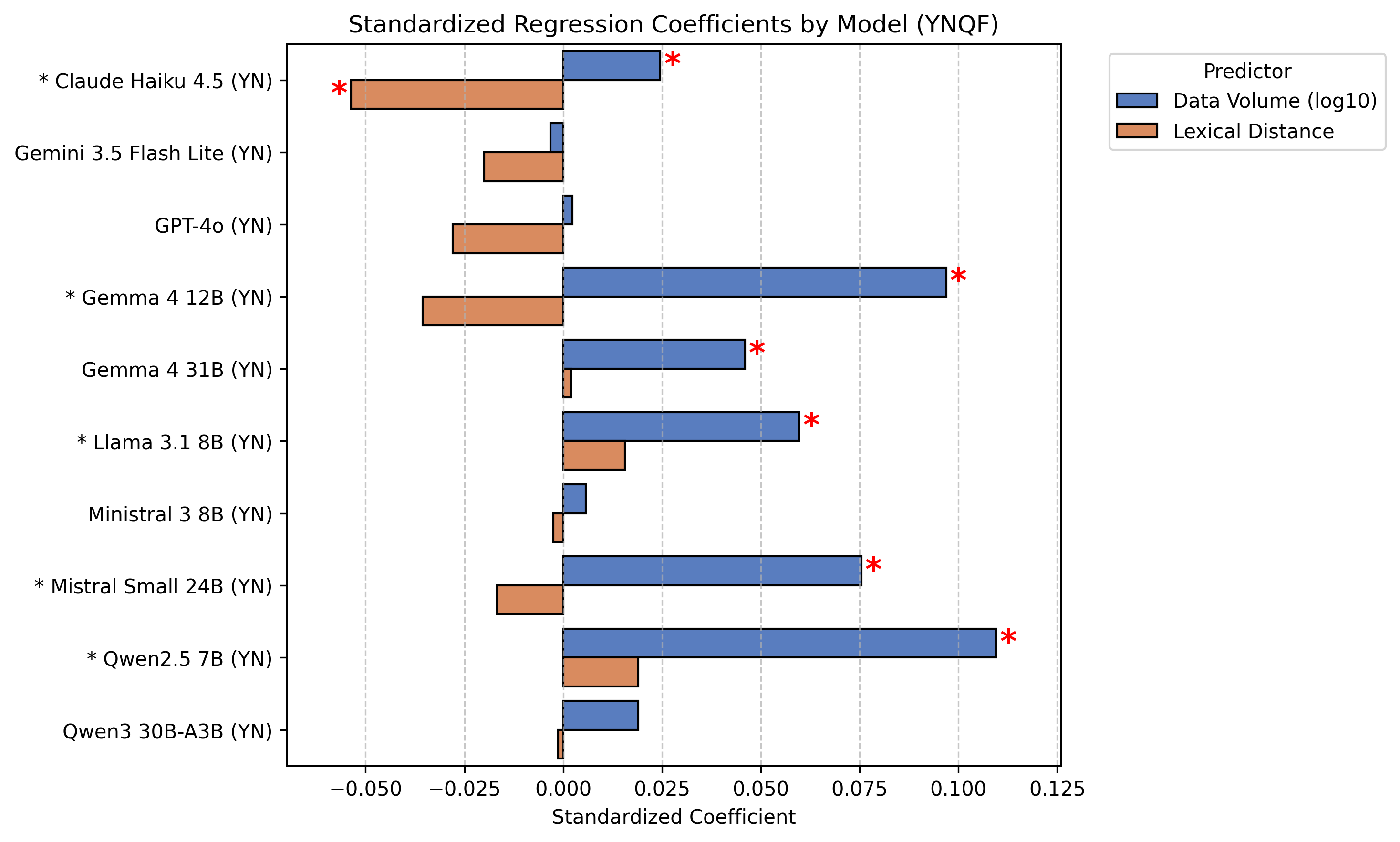}
  \caption{Standardized Regression Coefficients by Model (YNQF). An asterisk on a bar indicates a significant predictor, while an asterisk preceding a model name indicates a significant F-statistic.}
  \label{fig:regression_coefficients_YN}
\end{figure*}

\begin{figure*}[htb]
  \includegraphics[width=\textwidth]{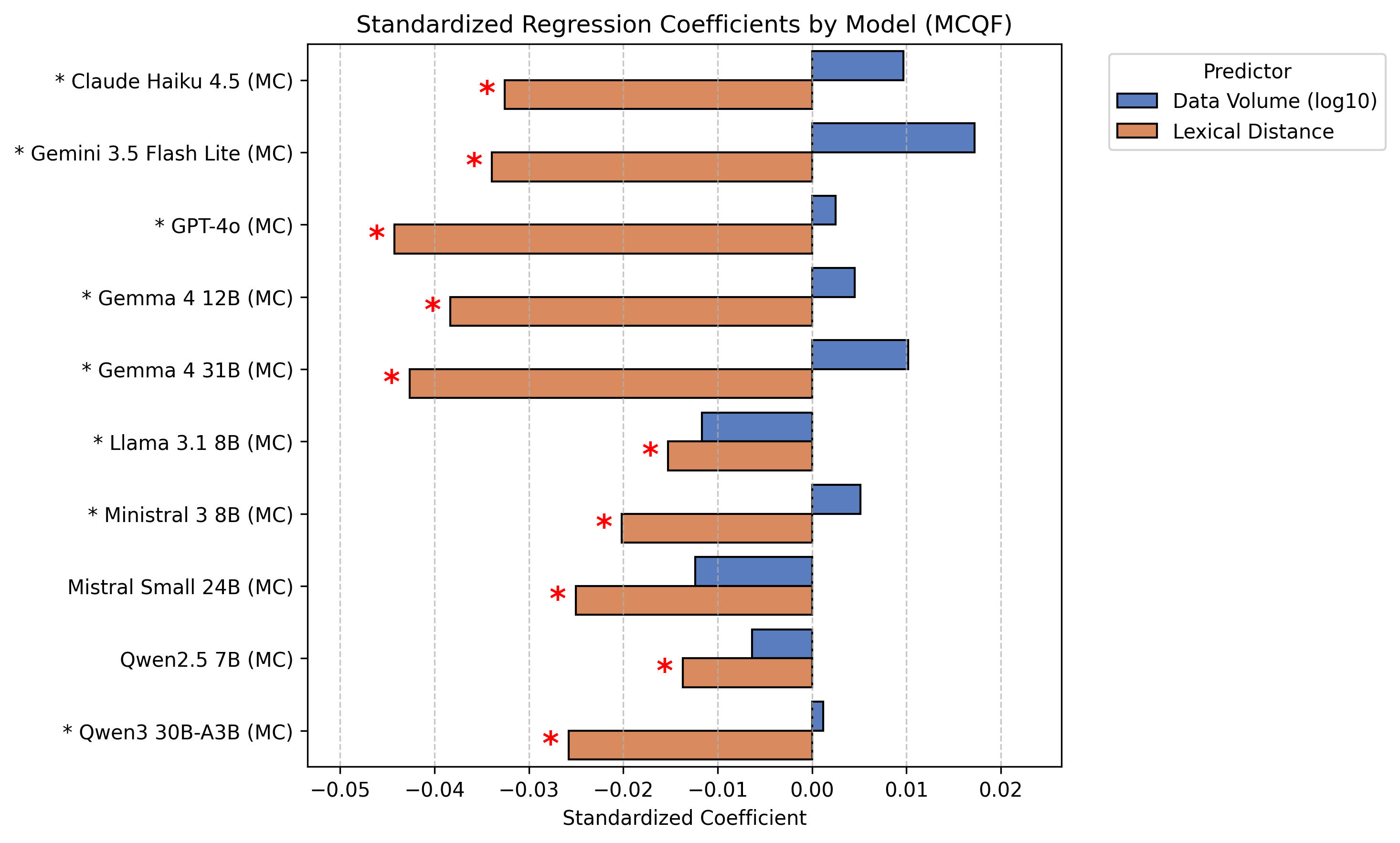}
  \caption{Standardized Regression Coefficients by Model (MCQF). An asterisk on a bar indicates a significant predictor, while an asterisk preceding a model name indicates a significant F-statistic.}
  \label{fig:regression_coefficients_MC}
\end{figure*}

\begin{figure*}[htb]
  \includegraphics[width=0.8\textwidth]{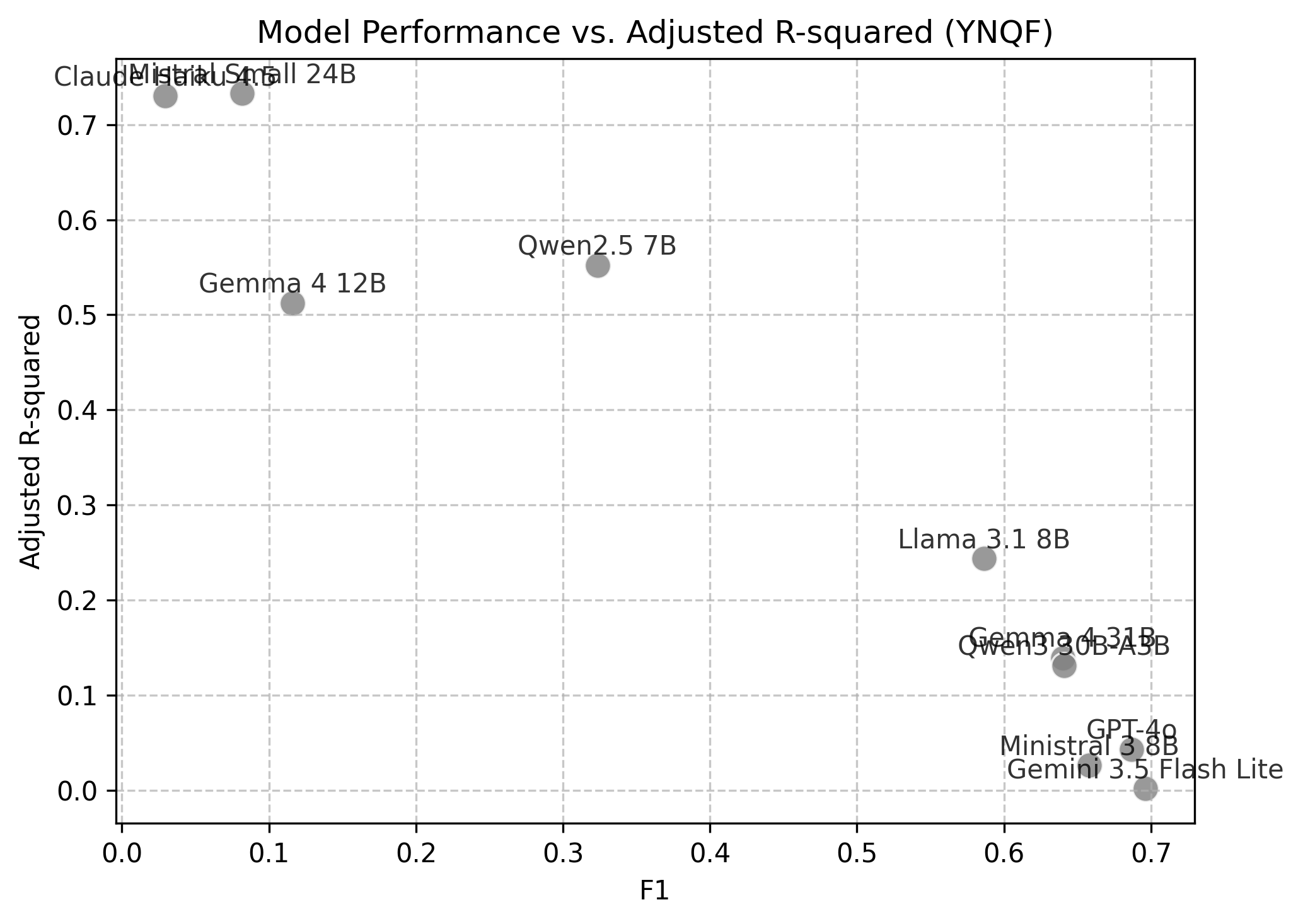}
  \caption{Model Performance vs. Adjusted R-squared (YNQF). Spearman's rank correlation $\rho = -0.964$ \quad ($p$ < 0.001)).}
  \label{fig:scatter_performance_adjR2_YN}
\end{figure*}

\begin{figure*}[htb]
  \includegraphics[width=0.8\textwidth]{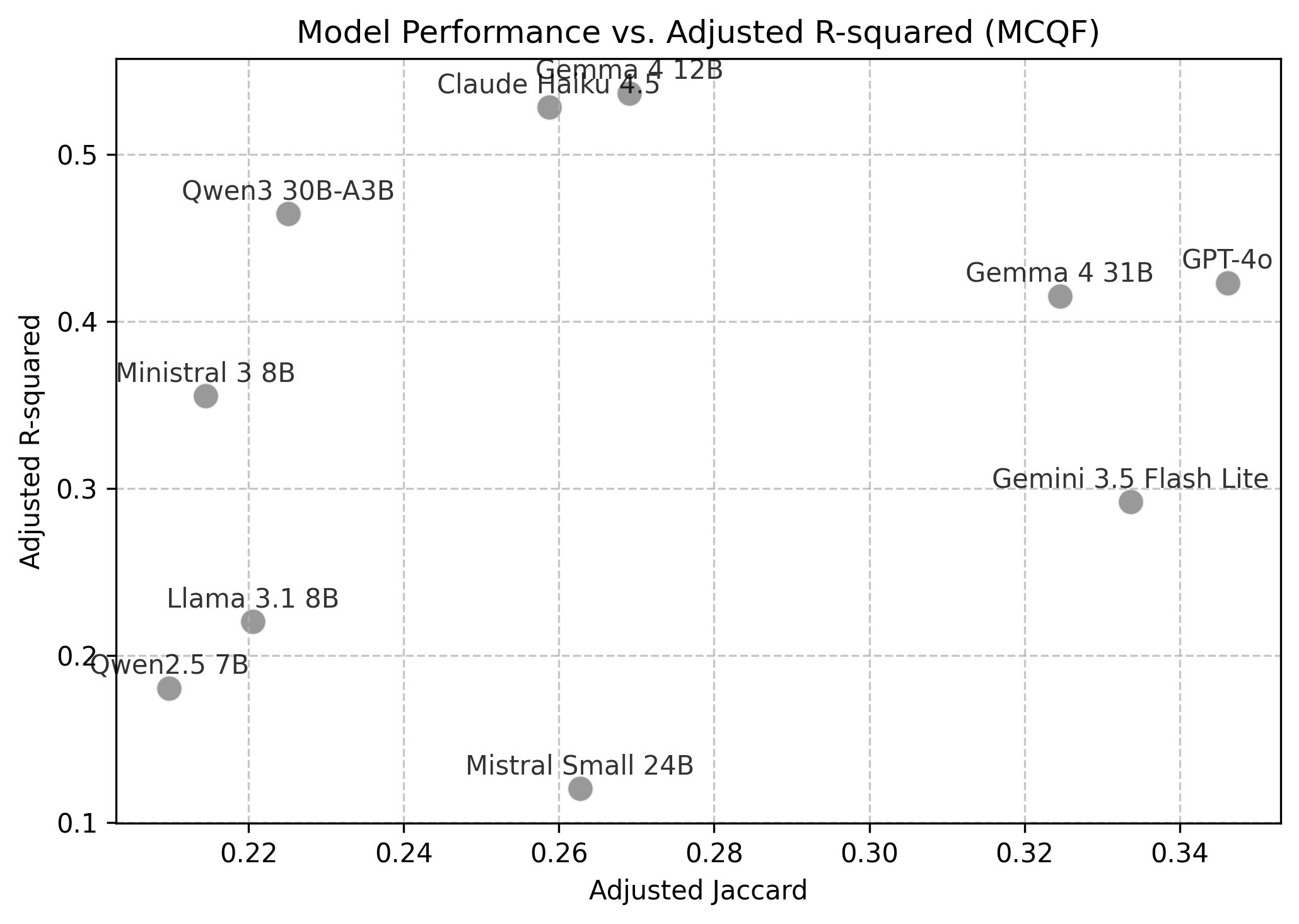}
  \caption{Model Performance vs. Adjusted R-squared (MCQF). Spearman's rank correlation $\rho = -0.309$ \quad ($p$ = 0.385)).}
  \label{fig:scatter_performance_adjR2_MC}
\end{figure*}

\end{document}